\documentclass[10pt,twocolumn,letterpaper]{article}

\usepackage{cvpr}              

\usepackage{graphicx}
\usepackage{amsmath}
\usepackage{amssymb}
\usepackage{booktabs}
\usepackage{multirow}
\usepackage[pagebackref,breaklinks,colorlinks]{hyperref}

\usepackage[capitalize]{cleveref}
\crefname{section}{Sec.}{Secs.}
\Crefname{section}{Section}{Sections}
\Crefname{table}{Table}{Tables}
\crefname{table}{Tab.}{Tabs.}

\def\cvprPaperID{3145 } 
\def\confName{CVPR}
\def\confYear{2023}

\begin{document}

\title{MD-VQA: Multi-Dimensional Quality Assessment for UGC Live Videos}

\author{Zicheng Zhang$^1$\thanks{These authors contributed equally to this work. The database is available at https://tianchi.aliyun.com/dataset/148818?t=1679581936815.}, Wei Wu$^2$\footnotemark[1], Wei Sun$^1$\footnotemark[1], Dangyang Tu$^1$, Wei Lu$^1$, \\ Xiongkuo Min$^1$, Ying Chen$^2$, Guangtao Zhai$^{1,3}$\thanks{Corresponding author.}\\
$^1$Institute of Image Communication and Network
Engineering, Shanghai Jiao Tong University\\
$^2$Alibaba Group\\
$^3$MoE Key Lab of Artificial Intelligence, AI Institute, Shanghai Jiao Tong University\\
\tt\small $^1$\{zzc1998,sunguwei,danyangtu,SJTU-Luwei,minxiongkuo,zhaiguangtao\}@sjtu.edu.cn, \\ \tt\small $^2$\{guokui.ww,yingchen\}@alibaba-inc.com
}
\maketitle


\begin{abstract}
User-generated content (UGC) live videos are often bothered by various distortions during capture procedures and thus exhibit diverse visual qualities. Such source videos are further compressed and transcoded by media server providers before being distributed to end-users. Because of the flourishing of UGC live videos, effective video quality assessment (VQA) tools are needed to monitor and perceptually optimize live streaming videos in the distributing process.  In this paper, we address \textbf{UGC Live VQA} problems by constructing a first-of-a-kind subjective UGC Live VQA database and developing an effective evaluation tool. Concretely, 418 source UGC videos are collected in real live streaming scenarios and 3,762 compressed ones at different bit rates are generated for the subsequent subjective VQA experiments. Based on the built database, we develop a \underline{M}ulti-\underline{D}imensional \underline{VQA} (\textbf{MD-VQA}) evaluator to measure the visual quality of UGC live videos from semantic, distortion, and motion aspects respectively. Extensive experimental results show that MD-VQA achieves state-of-the-art performance on both our UGC Live VQA database and existing compressed UGC VQA databases.
\end{abstract}

\section{Introduction}
\label{sec:intro}
\begin{figure}
    \centering
    \includegraphics[width=1\linewidth]{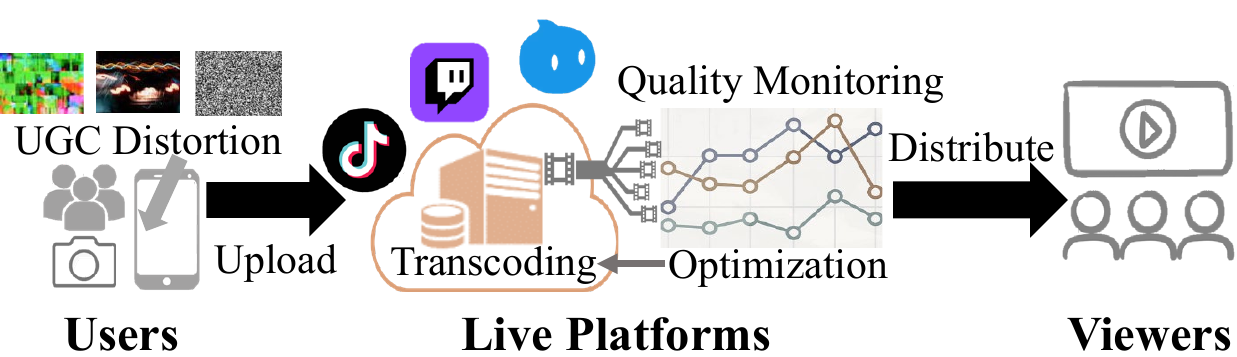}
    \caption{The distributing process of UGC live videos, where the users upload the videos degraded by the UGC distortions to the live platforms and the distorted videos are further compressed before being distributed to the viewers. The VQA models can monitor the quality changes of the compressed UGC live videos and adaptively optimize the transcoding setting.}
    \label{fig:process}
    \vspace{-0.4cm}
\end{figure}

With the rapid development of social media applications and the advancement of video shooting and processing technologies, more and more ordinary people are willing to tell their stories, share their experiences, and have their voice heard on social media or streaming media platforms such as Twitch, Tiktok, Taobao, etc. However, due to the lack of photography skills and professional equipment, the visual quality of user-generated content (UGC) videos may be degraded by in-the-wild distortions \cite{ying2021patch}. What's more, in common live platforms, live videos are encoded and distributed to end-users with very low delay, where compression algorithms have a significant influence on the visual quality of live videos because they can greatly reduce transmission bandwidth. As illustrated in Fig. \ref{fig:process}, video quality assessment (VQA) tools play an important role in monitoring, optimizing, and further improving the Quality of Experience (QoE) of end-users in UGC live streaming systems.

Currently, many UGC VQA databases have been carried out \cite{hosu2017konstanz,sinno2018large,wang2019youtube,li2020ugc,ying2021patch} to address the impact of general in-the-wild distortions on video quality, while some compression VQA databases \cite{seshadrinathan2010study,vu2014vis3,li2019avc} are proposed to study the influence of compression artifacts. Then some compressed UGC VQA databases \cite{li2020ugc,wang2021rich,icme2021} are further constructed to solve the problem of assessing the quality of UGC videos with compression distortions. However, they are either small in scale or employ high-quality UGC videos as the sources, and all of the mentioned databases lack videos in live streaming scenes. Therefore, there is a lack of a proper \textbf{UGC Live VQA} database to develop and validate the video quality measurement tools for live streaming systems.


\begin{table*}[t]\small
\renewcommand\arraystretch{.9}
\renewcommand\tabcolsep{4.5pt}
\centering
\caption{Review of common VQA databases, where 'UGC+Compression' refers to manually encoding the UGC videos with different compression settings.}
\vspace{-0.2cm}
\begin{tabular}{ccccccc}
\toprule
Database    &Year   & Duration/s & Ref. Num.& Scale   & Scope & Subjective Evaluating Format    \\
\midrule
CVD2014\cite{nuutinen2016cvd2014}     &2014   &10-25  &- &234 &In-capture & In-lab\\
LIVE-Qualcomm \cite{ghadiyaram2017capture} & 2016 & 15 &- & 208 & In-capture & In-lab\\
\hline
KoNViD-1k\cite{hosu2017konstanz} &2017 &8 &- &1,200 & In-the-wild & Crowdsourced \\
LIVE-VQC\cite{sinno2018large} &2018 &10 & - &585  & In-the-wild & Crowdsourced\\
YouTube-UGC\cite{wang2019youtube} &2019 &20 &-&1,500  & In-the-wild & Crowdsourced \\
LSVQ\cite{ying2021patch} &2021 &5-12 & - &39,075 & In-the-wild & Crowdsourced \\ \hline
UGC-VIDEO\cite{li2020ugc} &2019 &\textgreater 10 &50                &550 &UGC + Compression & In-lab\\
LIVE-WC\cite{yu2021predicting}&2020 & 10 & 55 & 275 & UGC + Compression & In-lab \\
YT-UGC$^+$(Subset)\cite{wang2021rich} &2021 &20 &189 &567  & UGC + Compression & In-lab \\
ICME2021\cite{icme2021} &2021 & - & 1,000 & 8,000 & UGC + Compression & In-lab \\
\bf{TaoLive(proposed)} & 2022 & 8 & 418 &3,762 & UGC + Compression &  In-lab\\
\bottomrule
\end{tabular}

\label{tab:compare}
\vspace{-0.4cm}
\end{table*}

To address UGC Live VQA problems, we first construct a large-scale database named \textbf{TaoLive}, consisting 418 source UGC videos from the TaoBao \cite{taolive} live streaming platform and the corresponding 3,762 compressed videos at various bit rates. Then we perform a subjective experiment in a well-controlled environment. Afterward, we propose a no-reference (NR) \underline{M}ulti-\underline{D}imensional \underline{VQA} (\textbf{MD-VQA}) model to measure the visual quality of UGC live videos in terms of semantic, distortion, and motion aspects. The semantic features are extracted by pretrained convolutional neural network (CNN) model; the distortion features are extracted by specific handcrafted image distortion descriptors (i.e. blur, noise, block effect, exposure, and colorfulness); and the motion features are extracted from video clips by pretrained 3D-CNN models. Compared with existing UGC VQA algorithms, MD-VQA measures visual quality from multiple dimensions, and these dimensions correspond to key factors affecting live video quality, which thereby has better interpretability and performance. The contributions of this paper are summarized as below:

\begin{itemize}
    \item {\bf{We build a large-scale UGC Live VQA database targeted at the compression artifacts on the UGC live videos.}} We collect 418 raw UGC live videos that are diverse in content, distortion, and quality. Then 8 encoding settings are used, which provides 3,762 compressed UGC live videos in total. 
    \item {\bf{We carry out a well-controlled in-lab subjective experiment.}} 44 participants are invited to participate in the subjective experiment and a total of 165,528 subjective annotations are gathered. 
    \item {\bf{A multi-dimensional NR-VQA model is proposed,}} using pretrained 2D-CNN, handcrafted distortion descriptors, and pretrained 3D-CNN for the semantic, distortion, and motion features extraction respectively. The extracted features are then spatio-temporally fused to obtain the video-level quality score. The extensive experimental results validate the effectiveness of the proposed method.
\end{itemize}

\begin{figure}[!tp]
    \centering
   \begin{subfigure}{0.49\linewidth}
    \includegraphics[width=1\linewidth,height =0.67\linewidth ]{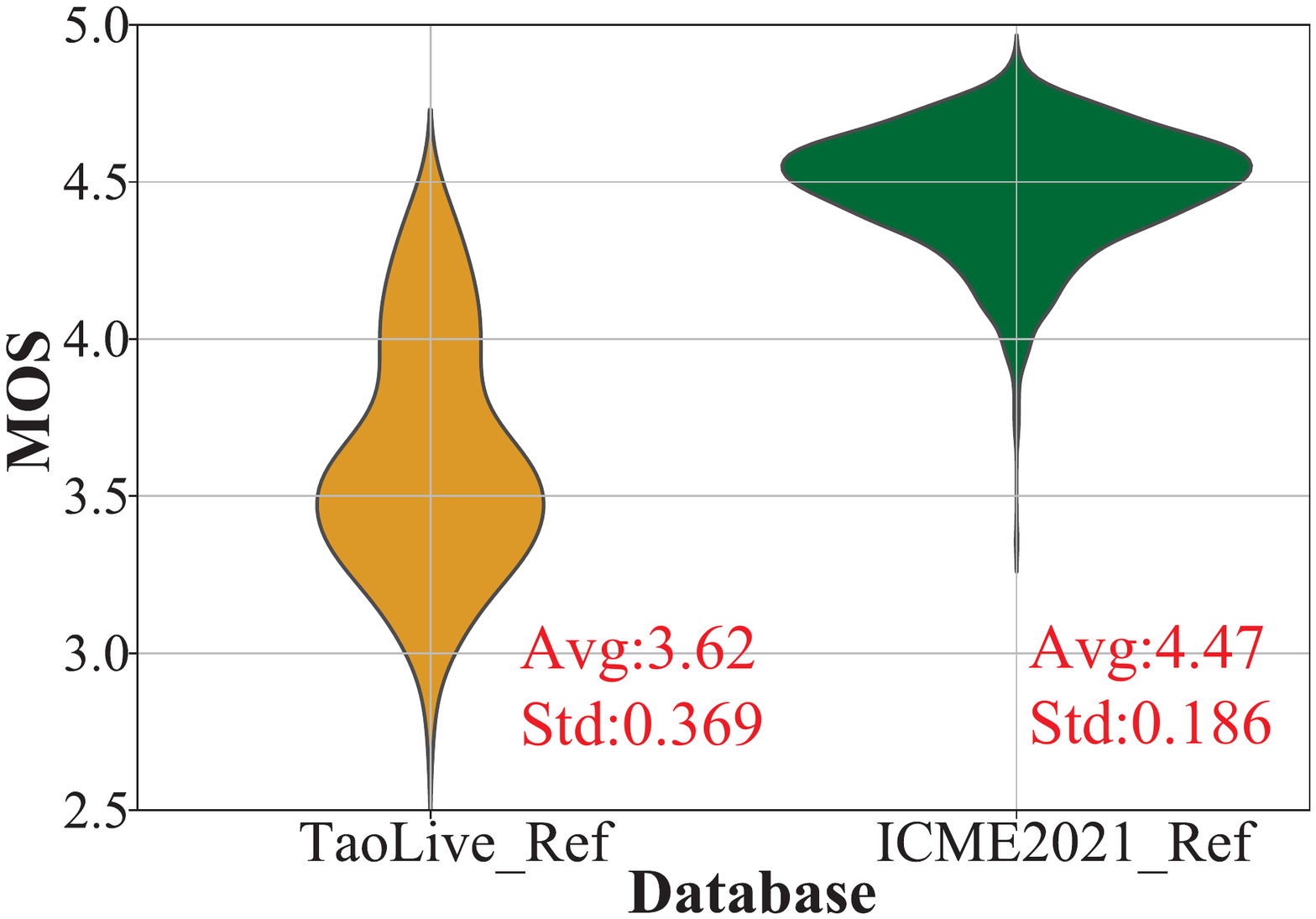}
    \caption{}
  \end{subfigure}
  \begin{subfigure}{0.49\linewidth}
    \includegraphics[width=1\linewidth,height =0.68\linewidth]{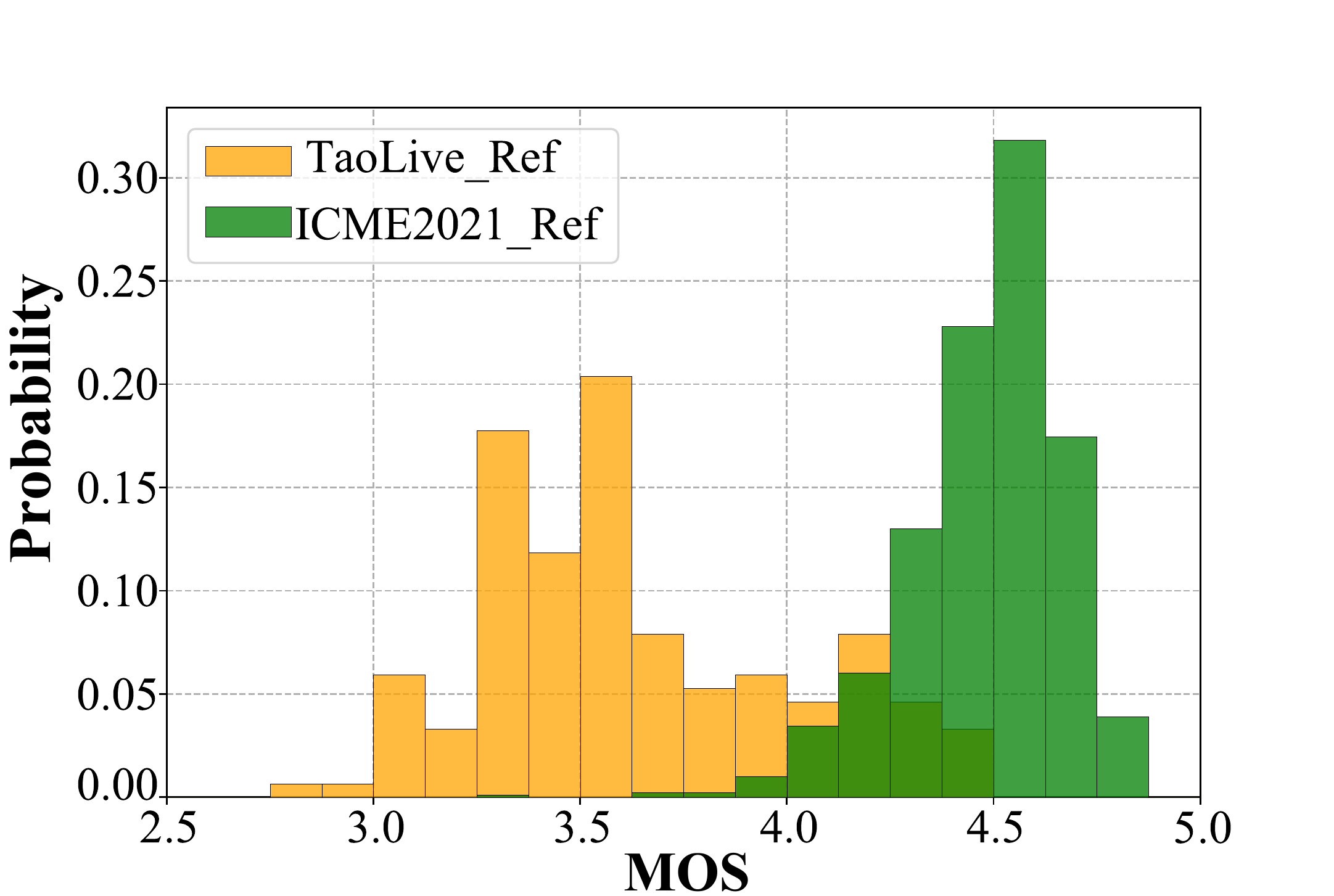}
    \caption{}
  \end{subfigure}
  \caption{Comparison of the quality distribution of reference videos between the TaoLive and ICME2021 \cite{icme2021} databases. The source UGC videos in the ICME2021 database are centered on high-quality levels while the quality levels of the source UGC videos in the TaoLive database are more diverse.}
  \label{fig:ref_compare}
  \vspace{-0.3cm}
\end{figure}

\section{Related Works}

\subsection{VQA Databases}
During the last decade, many VQA databases \cite{wang2016mcl,wang2017videoset,nuutinen2016cvd2014,ghadiyaram2017capture,hosu2017konstanz,sinno2018large,ying2021patch,yu2021predicting,wang2021rich} have been carried out to tackle the challenge of VQA problems and a review of the common VQA database is exhibited in Table \ref{tab:compare}. The early VQA databases \cite{wang2016mcl,wang2017videoset} usually collect limited numbers of source videos and manually introduce distortions such as compression and transmission error to generate distorted ones. Such databases are less diverse in content and distortions, which do not fit the scope of UGC videos. 
Then the CVD2014 \cite{nuutinen2016cvd2014}, LIVE-Qualcomm \cite{ghadiyaram2017capture}, and LIVE-VQC\cite{sinno2018large} databases are formed with videos that are captured by real cameras. However, the scale of the mentioned in-capture databases is relatively small and the included distortions are much simpler. 
Later, UGC VQA databases such as KoNViD-1k\cite{hosu2017konstanz}, YouTube-UGC\cite{wang2019youtube}, and LSVQ\cite{ying2021patch} gather in-the-wild UGC videos from online platforms, which have more diverse content and distortions and have significantly promoted the development of UGC VQA tasks. 

For UGC live videos, we can consider them as in-captured UGC videos followed by compressed distortions, where both in-the-wild distortions and compression distortions have a significant impact on the visual quality.
Although some works such as UGC-VIDEO\cite{li2020ugc}, LIVE-WC\cite{yu2021predicting}, and YT-UGC$^+$\cite{wang2021rich} attempt to assess the quality of UGC videos caused by common compression algorithms, the relatively small size of these databases makes it difficult to support the mainstream of data-driven models, e.g. deep learning-based models in Section \ref{ugc_vqa_models}. The recent ICME2021 \cite{icme2021} database is large in scale. However, as shown in Fig. \ref{fig:ref_compare}, source UGC videos in ICME2021 exhibit high visual quality, and thus can not reflect real source video quality distribution in live streaming systems. What's more, none of the mentioned databases includes source videos collected from practical live-streaming platforms.

\subsection{UGC VQA models}
\label{ugc_vqa_models}
\textbf{Handcrafted-based:} 
Handcrafted-based VQA models extract quality-aware features to model spatial and temporal distortions, such as natural scene statistics (NSS) features \cite{mittal2015completely,saad2014blind,liao2022exploring}, artifacts\cite{korhonen2019two,tu2021ugc,madhusudana2021st}, motion\cite{korhonen2019two,tu2021ugc,ebenezer2021chipqa}, etc.
For example, 
VIIDEO \cite{mittal2015completely} gives the intrinsic statistical regularities gathered from natural videos and assesses the video quality according to the regularities. V-BLIINDS \cite{saad2014blind} evaluates the video quality by using a spatio-temporal NSS model and a motion representation model. TLVQM \cite{korhonen2019two} computes low complexity and high complexity quality-aware features in two steps to obtain the final video quality. VIDEVAL \cite{tu2021ugc} carefully chooses representative quality-aware features among the mainstream NR-VQA models and regresses the representative features into quality values.

\textbf{Deep learning-based:} 
Considering the huge parameters of deep neural networks (DNN) and the relatively small scale of VQA databases, some VQA methods use pretrained DNN models for feature extraction. VSFA \cite{li2019quality} extracts deep semantic features with a pre-trained DNN model and learns the quality-aware representation with gated recurrent units (GRUs). 
To enhance the understanding of video motion features, some studies \cite{ying2021patch,li2022blindly,sun2022deep} further attempt to extract motion features with 3D-CNN models pre-trained on the video action recognition databases to help detect video motion distortions and have yielded good performance. Later, some Transformer-based VQA methods are carried out. LSCT \cite{you2021long} first extracts frame-wise perceptual quality features and then feeds the features into a long short-term convolutional Transformer to predict the video quality. FAST/FASTER-VQA \cite{wu2022fast,wu2022neighbourhood} proposes Grid Mini-patch Sampling and forms the video sampling results as fragments, which are put into a fragment-modified video swin transformer \cite{liu2022video} for video quality representation.

\begin{figure}
    \centering
    \includegraphics[width = 1\linewidth]{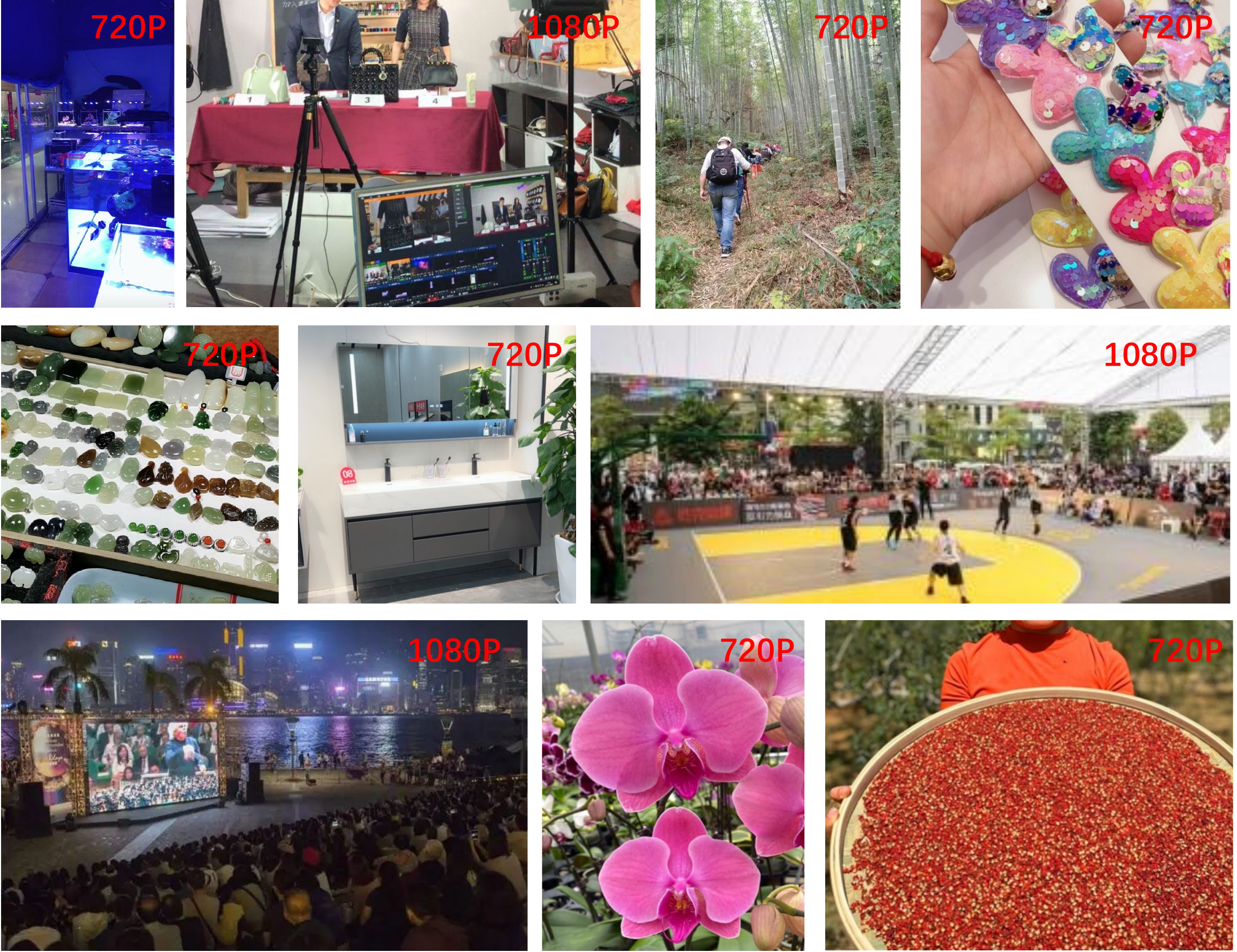}
    \caption{Sample frames of the videos in the proposed TaoLive Database, where the resolutions are marked on the top right. Additionally, the frame samples are cropped to delete some sensitive content such as human faces and watermarks for exhibition.}
    \label{fig:samples}
    \vspace{-0.5cm}
\end{figure}

\section{UGC Live Dataset and Human Study}
\subsection{Videos Collection}
As illustrated in Fig. \ref{fig:process}, users upload their live video streams to platforms, and platforms compress the video streaming at different bit rates and distribute one of them to viewers according to the Internet bandwidth or the viewers' choice. To reflect the real quality distribution of in-captured live video, we first collect large-scale uncompressed raw UGC videos from the Taolive \cite{taolive} platform, a very popular live platform in China. Then, we manually select raw UGC videos that contain the scenes of technology, fashion, food, daily life, financial sales, etc, to ensure content diversity. 
In the last, we collect 418 raw UGC videos (110 videos have resolutions of 720P while 318 videos have resolutions of 1080P), and each raw UGC video is cropped into about 8s duration as source UGC live videos.

We use the open-source compression tools, \textit{FFmpeg}\cite{ffmpeg}, to compress source UGC live videos by 8 Constant Rate Factors (CRF) of H.265 including 16, 20, 24, 28, 32, 36, 40, and 44 to close to the distributing process of live platforms. Therefore, 3,344=418$\times$8 compressed UGC videos are generated and a total of 3,762 = 3,344 + 418 UGC videos are collected for evaluation, the samples of which are exhibited in Fig. \ref{fig:samples}.

\subsection{Human Study}
The human study is carried out in a well-controlled environment. 44 humans including 20 males and 24 females are invited to participate in the subjective experiment. The viewers are seated about 1.5 times the screen height (45cm) with normal indoor illumination and the videos are played on an iMac monitor which supports a resolution up to 4096$\times$2304. Before the viewers start to evaluate the UGC live videos, a short introduction is given to get the viewers familiar with the equipment and quality assessment tasks. We split the experiment into 76 small sessions and each session contains 50 UGC live videos with no content overlap. The viewers participate in all the sessions individually and each session lasts about 30 minutes. There is at least 1-hour break between the sessions and each subject is allowed to attend no more than 2 sessions in a single day. During the sessions, each video is played only once and the viewers can rate the video quality from 1 to 5, with a minimum interval of 0.1. We make sure that each UGC live video is evaluated by the 44 invited viewers and 165,528=3,762$\times$44 subjective ratings are collected in the end.

\subsection{Subjective Data Analysis}
According to the recommendation of ITU-R BT.500-13 \cite{bt2002methodology}, we compute the z-scores as the quality label of UGC live videos:

\begin{equation}
z_{i j}=\frac{r_{i j}-\mu_{i}}{\sigma_{i}},
\end{equation}
where $r_{ij}$ represents the quality rating given by the $i$-th subject on the $j$-th UGC live video, $\mu_{i}=\frac{1}{N_{i}} \sum_{j=1}^{N_{i}} r_{i j}$, $\sigma_{i}=\sqrt{\frac{1}{N_{i}-1} \sum_{j=1}^{N_{i}}\left(r_{i j}-\mu_{i}\right)}$, and $N_i$ is the number of UGC live videos evaluated by subject $i$.
Then we remove the quality labels from unreliable subjects according to the recommended subject rejection procedure in \cite{bt2002methodology}. Finally, the z-scores are linearly rescaled to $[1,5]$ and the mean opinion score (MOS) of the UGC video $j$ is obtained by averaging the rescaled z-scores: 

\begin{equation}
M O S_{j}=\frac{1}{M} \sum_{i=1}^{M} z_{i j}^{'},
\end{equation}
where $M O S_{j}$ represents the MOS for the $j$-th UGC video, $M$ is the number of the valid subjects, and $z_{i j}^{'}$ are the rescaled z-scores.

\begin{figure}
  \centering
  \begin{subfigure}{0.93\linewidth}
    
    \includegraphics[width=1\linewidth]{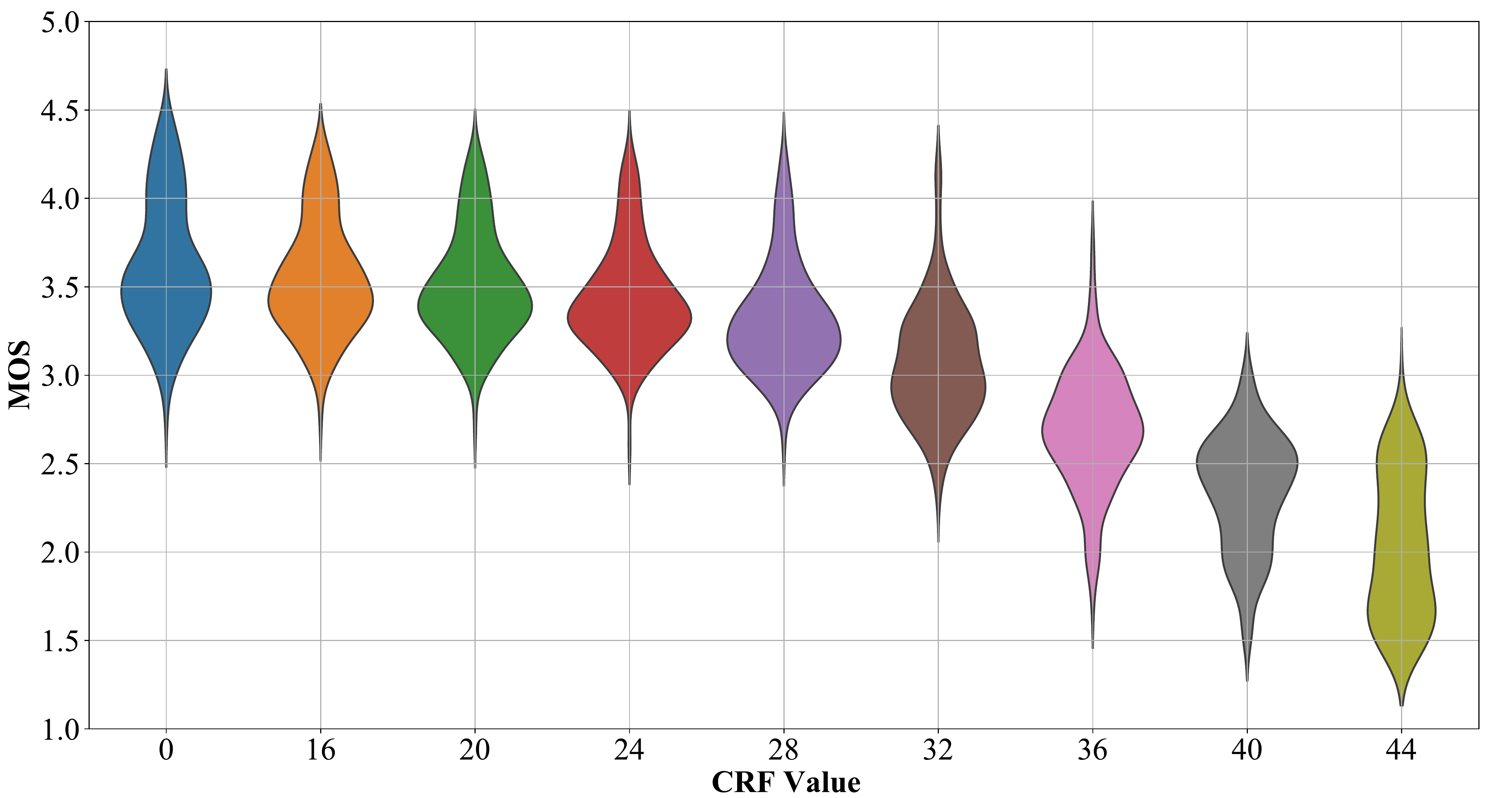}
    \caption{MOS distributions for different CRF values, where `0' indicates source videos.}
    \label{fig:taolive_dist:a}
  \end{subfigure}
  \\
  
  \begin{subfigure}{0.45\linewidth}
    \includegraphics[width=1\linewidth,height =0.63\linewidth ]{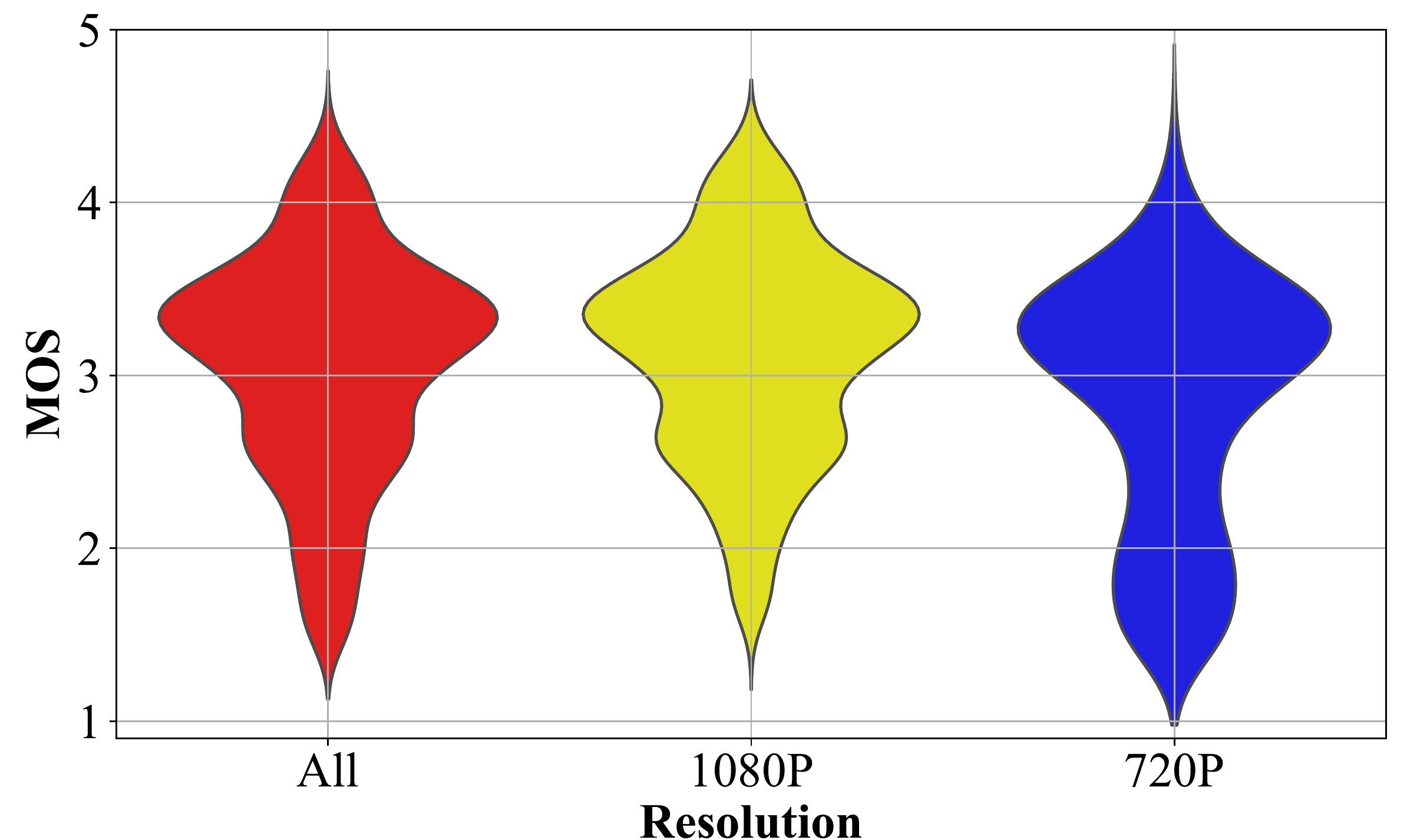}
    \caption{MOS distributions for different resolutions.}
    \label{fig:taolive_dist:b}
  \end{subfigure}
  \begin{subfigure}{0.45\linewidth}
    \includegraphics[width=1\linewidth]{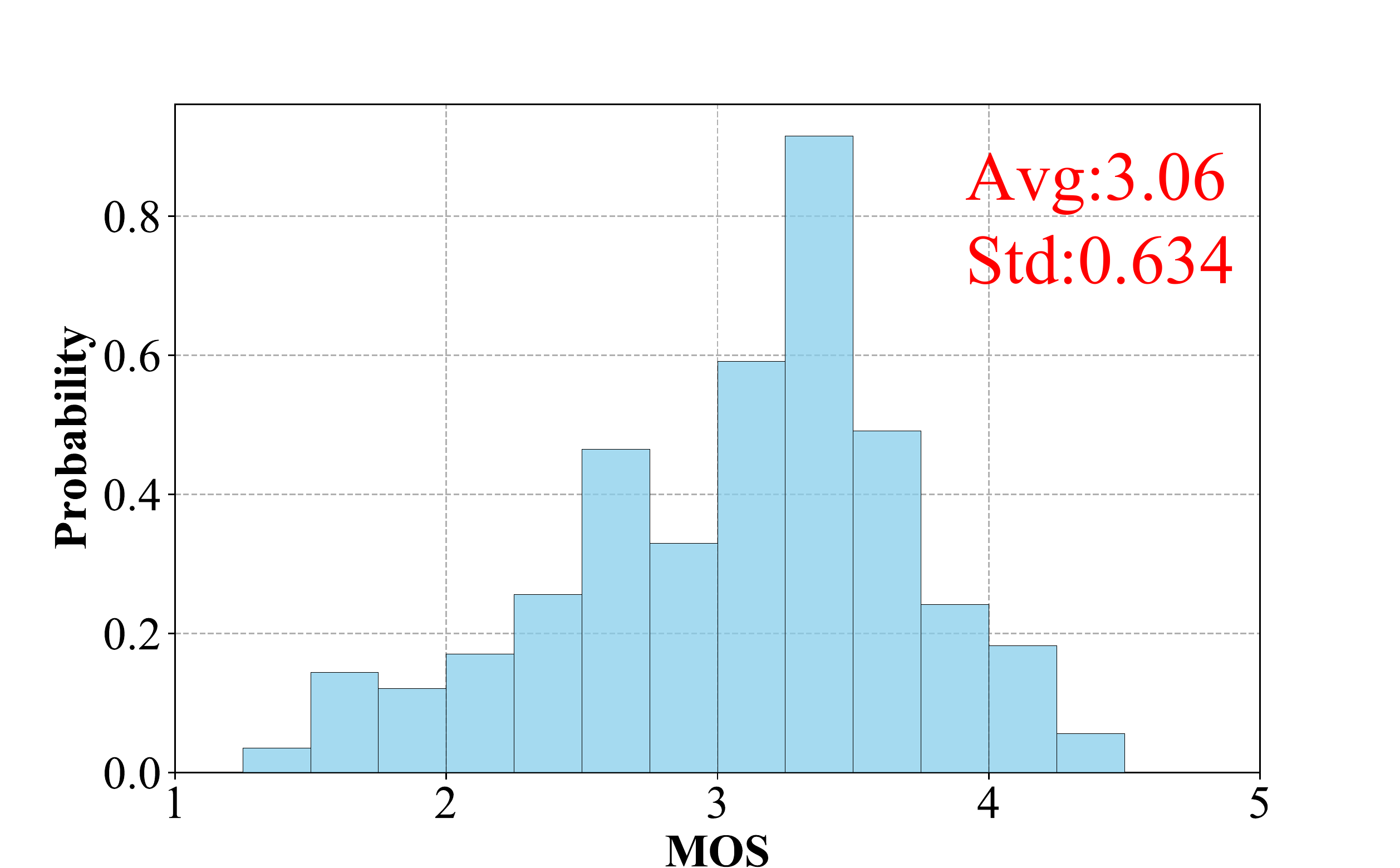}
    \caption{Detailed MOS distribution for the TaoLive database.}
    \label{fig:taolive_dist:c}
  \end{subfigure}
  
  \caption{Illustration of the proposed TaoLive database's MOS distributions from different perspectives.}
  \label{fig:taolive_dist}
  \vspace{0.3cm}
\end{figure}

We further plot the MOS distributions from the CRF and resolution perspectives. As shown in Fig. \ref{fig:taolive_dist:a}, conservative CRF parameter selection (16$\sim$24) introduces slight perceptual quality loss to the UGC live videos. When CRF increases from 28 to 44, the downward trend of perceptual quality is more obvious. Moreover, when CRF$\geq$40, nearly no compressed UGC video gains higher quality score than 3, which suggests that the 40+ CRF selection can result in a viewing experience below average. Such phenomena can provide useful guidelines for the compression strategy of live platforms. From Fig. \ref{fig:taolive_dist:b}, we can find that the general quality of UGC videos with a resolution of 720P is lower than the UGC videos with a resolution of 1080P, which fits the common sense that lower resolutions lead to poorer visual quality levels.

\begin{figure*}
    \centering
    \includegraphics[width=0.9\linewidth]{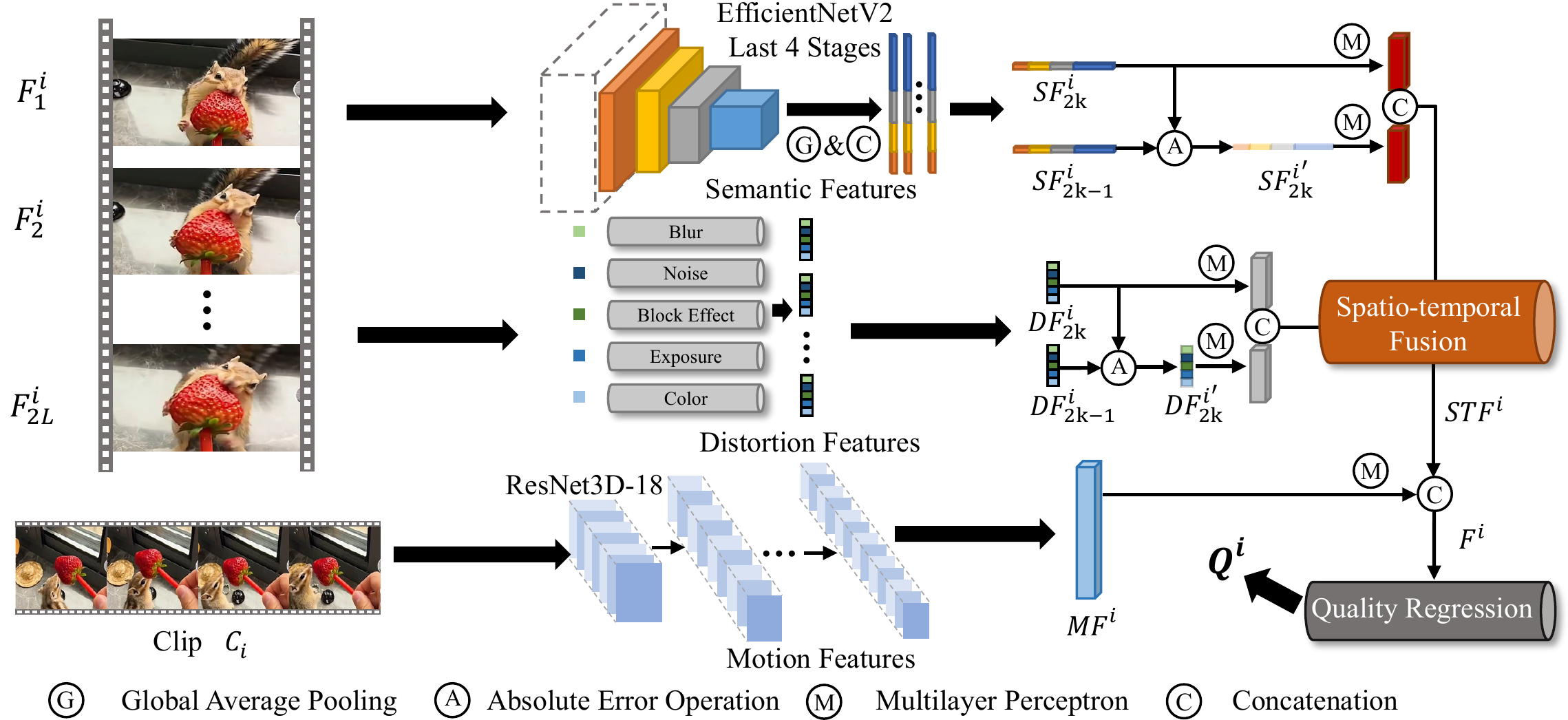}
    \caption{The framework of the proposed method, where the semantic, distortion, and motion features are extracted by the pretrained EfficientNetV2 \cite{tan2021efficientnetv2}, handcrafted distortion descriptors, and pretrained ResNet3D-18 \cite{hara3dcnns} respectively. The absolute error between the adjacent frames' semantic and distortion features is used to reflect the temporal quality fluctuations. Finally, the multi-dimensional features are spatio-temporally fused and regressed into quality values. }
    \label{fig:framework}
    \vspace{-0.3cm}
\end{figure*}

\section{Proposed Method}
\label{sec:method}
The framework of the proposed MD-VQA model is illustrated in Fig. \ref{fig:framework}, which includes the feature extraction module, feature fusion module, and feature regression module. Specifically, quality-aware features are extracted from multiple dimensions including the semantic, distortion, and motion aspects. What's more, the feature error between adjacent frames is employed to reflect the temporal quality fluctuation. Then the obtained multi-dimensional features are fused in spatio-temporal manners and mapped to quality scores via the quality regression module.

\subsection{Feature Extraction}

\label{sec:pre}
Given a video whose number of frames and frame rate is $n$ and $r$, we split the video into $\frac{n}{r}$ clips for feature extraction and each clip lasts for 1s.
For each clip $C_{i}$ ($i$ represents the index of the clip), $2L$ frames are uniformly sampled for semantic and distortion feature extraction while the whole clip is employed for motion feature extraction.

\subsubsection{Semantic Feature Extraction}
Different semantic contents shall have diverse impacts on humans' tolerance for different distortions \cite{li2019quality}. For example, humans are more able to endure blur distortions on flat and texture-less objects such as clear sky and smooth walls \cite{zhang2021full,zhang2023perceptual}. However, the blur distortions can be unacceptable on objects that are rich in texture such as rough rocks and complex plants. It is also believed that semantic information can help identify the existence and extent of the perceived distortions \cite{dodge2016understanding}. Moreover, it has been proven that human perception is highly affected by compression \cite{jayant1993signal,xu2020state}. Thus it is quite reasonable to incorporate the semantic information into the compressed UGC video quality assessment.

Considering that visual perception is a hierarchical structure, that is, input visual information is perceived hierarchically from low-level features to high-level \cite{lu2022cnn,wang2021multi,wang2022subjective}, we propose to utilize multi-scale features extracted from the last 4 stages of the pretrained EfficientNetV2 \cite{tan2021efficientnetv2} as the frame-level semantic information:
\begin{equation}
\begin{aligned}
     SF_{l}^{i} &= \alpha_{1}\oplus \alpha_{2} \oplus \alpha_{3} \oplus \alpha_{4}, l \in \{1,\cdots,2L\}, \\
      \alpha_{j} &= {\rm{GAP}}(L_{j}(F_{l}^{i})), j \in \{1,2,3,4\},
\end{aligned} 
\end{equation}
where $SF_{l}^{i}$ indicates the extracted semantic features from the $l$-th sampled frame $F_{l}^{i}$ of clip $C_{i}$, $\oplus(\cdot)$ stands for the concatenation operation, $\rm{GAP}(\cdot)$ represents the global average pooling operation, $L_{j}(F_{l}^{i})$ stands for the feature maps obtained from $j$-th last layer of EfficientNetV2, and $\alpha_{j}$ denotes the average pooled features from $L_{j}(F_{l}^{i})$. 

\subsubsection{Distortion Feature Extraction}
Various types of distortions exist in UGC videos and only utilizing semantic information is insufficient to model the distortion perception of UGC videos. What's more, the original UGC distortions can exhibit dissimilar quality representations under different levels of compression. For example, the blur is less sensitive to compression \cite{zhang2021no,zhang2022nofre} since compression usually wipes out high-frequency information, and noise can be eased or even disappear when higher compression levels are applied \cite{al1998lossy}.  Therefore, to better improve the quality representation of the proposed method, some handcrafted distortion descriptors are further employed for quality prediction, which include blur \cite{zhan2017no}, noise \cite{chen2015efficient}, block effect \cite{wang2002no}, exposure \cite{korhonen2019two}, and colorfulness \cite{panetta2013no}. Then the frame-level distortion features can be derived as:
\begin{equation}
    DF_{l}^{i} = \Psi(F_{l}^{i}), l \in \{1,\cdots,2L\}, 
\end{equation}
where $DF_{l}^{i}$ represents the extracted distortion features from the $l$-th sampled frame $F_{l}^{i}$ of clip $C_{i}$ and $\Psi(\cdot)$ stands for the distortion feature extraction process.

\subsubsection{Motion Feature Extraction}
UGC live videos are often bothered with motion distortions resulting from the unstable shooting environment as well as the restricted bit rates. However, these motion distortions such as the video shaking and the motion blur are difficult to recognize from the spatial features alone. Furthermore, video compression deeply depends on motion estimation \cite{lu2019dvc,furht2012motion}, which indicates that motion distortions can influence the quality of video compression.  Therefore, to help the model better understand the motion information, we propose to use the pretrained 3D-CNN backbone, ResNet3D-18 \cite{hara3dcnns}, to capture clip-level motion distortions:
\begin{equation}
    MF^{i} = \Gamma(C_{i}),
\end{equation}
where $MF^{i}$ denotes the motion features extracted from clip $C_{i}$ and $\Gamma(\cdot)$ represents the motion feature extraction operation.

To sum up, given the $i$-th clip $C_{i}$ of the video, we can obtain the clip-level semantic features $SF^{i} \in \mathbb{R}^{2L \times N_{S}}$, the distortion features $DF^{i} \in \mathbb{R}^{2L \times N_{D}}$, and the motion features $MF^{i} \in \mathbb{R}^{1 \times N_{M}}$, where $N_{S}$, $N_{D}$, and $N_{M}$ represent the number of channels for the semantic, distortion, and motion features respectively.

\subsection{Feature Fusion}
It has been proven in \cite{narwaria2012low} that videos with better quality tend to have smaller quality fluctuations while videos with lower quality tend to have larger quality fluctuations. Therefore, to quantify the fluctuations that are highly correlated with human perception, we propose to employ the absolute error between adjacent semantic and distortion features for temporal quality fluctuations reflection:
\begin{equation}
\begin{aligned}
    SF_{2k}^{i'} &= |SF_{2k}^{i}-SF_{2k-1}^{i}|,k \in \{1,\cdots,L\}, \\
    DF_{2k}^{i'} &= |DF_{2k}^{i}-DF_{2k-1}^{i}|,k \in \{1,\cdots,L\},
\end{aligned}
\end{equation}
where $SF_{2k}^{i'}$ and $DF_{2k}^{i'}$ represent the absolute error between adjacent semantic and distortion features. Then the spatio-temporal fusion can be derived as:
\begin{equation}
\begin{aligned}
    SD_{2k}^{i} = \omega(SF_{2k}^{i}) & \oplus \omega(DF_{2k}^{i}) \oplus \omega(SF_{2k}^{i'}) \oplus \omega(DF_{2k}^{i'}), \\
     & STF^{i}  =  W_{L}^{1}  (SD^{i^{T}}),
\end{aligned}
\end{equation}
where $\oplus(\cdot)$ stands for the concatenation operation, $\omega(\cdot)$ represents the learnable Multilayer Perceptron (MLP), $SD_{2k}^{i} \in \mathbb{R}^{1 \times N_{SD}}$ indicates the frame-level spatial-fused features obtained from semantic and distortion features, $SD^{i^{T}} \in \mathbb{R}^{N_{SD} \times L}$  is the transposition result of the clip-level semantic and distortion features $SD^{i} \in \mathbb{R}^{L\times N_{SD}}$, $W_{L}^{1}$ is a learnable linear mapping operation to fuse the $SD^{i^{T}}$ in the temporal domain, and we finally obtain the spatio-temporal fused features $STF^{i} \in \mathbb{R}^{N_{SD}\times1}$. To further introduce the quality-aware motion features, we concatenate the spatio-temporal features with the motion features:
\begin{equation}
    F^{i} = STF^{i^{T}} \oplus \omega(MF^{i}),
\end{equation}
where $STF^{i^{T}} \in \mathbb{R}^{1\times N_{SD}}$, the final clip-level quality-aware representation $F^{i} \in \mathbb{R}^{1\times (N_{SD}+N_{M}')}$ and $N_{M}'$ is adjusted number of channels for the motion features after MLP operation.

\subsection{Feature Regression}
After the feature extraction process described above, we use the three-stage fully-connected layers to regress the clip-level quality-aware representation $F^{i}$ into quality values:
\begin{equation}
    Q^{i} = \mathbf{FC}(F^{i}),
\end{equation}
where $\mathbf{FC}(\cdot)$ indicates the fully-connected layers and $Q_{i}$ stands for the quality value of clip $C_{i}$.
Consequently, the overall UGC live video quality can be obtained via average pooling:
\begin{equation}
    \boldsymbol{Q} = 1/\frac{n}{r} \sum_{1}^{\frac{n}{r}} Q^{i},
\end{equation}
where $\boldsymbol{Q}$ is the video quality value and $\frac{n}{r}$ represents the number of clips.
We simply use the Mean Squared Error (MSE) as the loss function:
\begin{equation}
    Loss = \frac{1}{n}\sum_{m=1}^{n}\left(Q'_{m}-Q_{m}\right)^{2}
\end{equation}
where $n$ indicates the number of videos in a mini-batch, $Q'_{m}$ and $Q_{m}$ are the subjective quality labels and predicted quality levels respectively.

\begin{table*}[!th]\small
\centering
\caption{Experimental performance on the compressed UGC VQA databases. `Hand' indicates using handcrafted-based features while `Deep' indicates using deep learning-based features. Best in {\bf\textcolor{red}{red}} and second in {\bf\textcolor{blue}{blue}}. }
\begin{tabular}{l|cc|cc|cc|cc}
\toprule
 \multirow{2}{*}{Method} & \multirow{2}{*}{Hand} &\multirow{2}{*}{Deep} & \multicolumn{2}{c|}{LIVE-WC} & \multicolumn{2}{c|}{YT-UGC$^+$}  & \multicolumn{2}{c}{TaoLive}\\  \cline{4-9}
   & & & SRCC$\uparrow$       & PLCC$\uparrow$      & SRCC$\uparrow$       & PLCC$\uparrow$ & SRCC$\uparrow$       & PLCC$\uparrow$  \\ \hline

BRISQUE (TIP, 2012)\cite{brisque} & \checkmark&$\times$      & 0.787 & 0.788   & 0.303 & 0.309  &0.771 &0.777  \\

TLVQM (TIP, 2019)\cite{korhonen2019two} & \checkmark&$\times$& 0.838 & 0.830   & 0.672 & 0.697    &0.862 &0.869        \\
VIDEVAL (TIP, 2021) \cite{tu2021ugc}  &\checkmark&$\times$  & 0.812 & 0.825   & 0.660 & 0.662    &0.914 &0.910     \\ \hline
VSFA (ACM MM, 2019) \cite{li2019quality} & $\times$&\checkmark   & 0.856 & 0.857   & 0.784 & 0.783  &0.920  &0.917  \\
PVQ (CVPR, 2021) \cite{ying2021patch} & $\times$&\checkmark &0.901 & {0.909}  &0.775 &0.776 &0.916 & 0.919  \\
BVQA (TCSVT, 2022)\cite{li2022blindly} & $\times$&\checkmark & 0.912 &0.916  &0.777 &0.781  &0.926 & 0.922 \\
SimpleVQA (ACM MM, 2022)\cite{sun2022deep} &$\times$&\checkmark & \bf\textcolor{blue}{0.927} & \bf\textcolor{blue}{0.920}   & \bf\textcolor{blue}{0.789} & \bf\textcolor{blue}{0.784}   & \bf\textcolor{blue}{0.932} & \bf\textcolor{blue}{0.926} \\ \hline

\bf{MD-VQA}(Ours)  & \checkmark&\checkmark  &\bf\textcolor{red}{0.931} & \bf\textcolor{red}{0.937} & \bf\textcolor{red}{0.822} & \bf\textcolor{red}{0.828}  &\bf\textcolor{red}{0.942} &\bf\textcolor{red}{0.945}       \\          \bottomrule
\end{tabular}
\label{tab:intra}
\vspace{-0.2cm}
\end{table*}

\begin{table}[!th]\small
\centering
\caption{Experimental performance of the ablation study, where SF, DF, and MF indicate the semantic features, distortion features, and motion features respectively. }
\setlength{\tabcolsep}{2pt}
\begin{tabular}{c|cc|cc|cc}
\toprule
\multirow{2}{*}{Feature} & \multicolumn{2}{c|}{LIVE-WC} & \multicolumn{2}{c|}{YT-UGC$^+$}& \multicolumn{2}{c}{TaoLive} \\  \cline{2-7}
& SRCC$\uparrow$       & PLCC$\uparrow$    & SRCC$\uparrow$       & PLCC$\uparrow$  & SRCC$\uparrow$       & PLCC$\uparrow$ \\\hline

SF  &0.911 &0.910  &0.785 &0.787  &0.931 & 0.934\\
DF  &0.640 &0.658  &0.277 &0.381  &0.603 & 0.638  \\
MF  &0.841 &0.858  &0.537 &0.561  &0.909 & 0.912    \\
SF+DF   &\bf\textcolor{blue}{0.925} &{0.922} &\bf\textcolor{blue}{0.805} &\bf\textcolor{blue}{0.824}  &0.935 &0.936        \\
SF+MF   &0.921 & \bf\textcolor{blue}{0.924}  &0.792 &0.789 &\bf\textcolor{blue}{0.940} & \bf\textcolor{blue}{0.941}    \\
DF+MF   &0.857 &0.861  &0.573 &0.631 &0.925 &0.926      \\ \hline
All &\bf\textcolor{red}{0.931} & \bf\textcolor{red}{0.937} & \bf\textcolor{red}{0.822} & \bf\textcolor{red}{0.828} &\bf\textcolor{red}{0.942} &\bf\textcolor{red}{0.945} \\  \bottomrule
\end{tabular}
\vspace{-0.35cm}
\label{tab:ablation}
\end{table}

\begin{table}[!th]\small
\centering
\caption{Ablation study results for absolute error (ABS) and feature fusion module (FFM), where ABS is replaced with error and FFM is replaced with concatenation. }
\setlength{\tabcolsep}{1pt}
\begin{tabular}{c|cc|cc|cc}
\toprule
\multirow{2}{*}{Model} & \multicolumn{2}{c|}{LIVE-WC} & \multicolumn{2}{c|}{YT-UGC$^+$}& \multicolumn{2}{c}{TaoLive} \\  \cline{2-7}
& SRCC$\uparrow$       & PLCC$\uparrow$    & SRCC$\uparrow$       & PLCC$\uparrow$  & SRCC$\uparrow$       & PLCC$\uparrow$ \\\hline

w/o ABS   & 0.912 &\bf\textcolor{blue}{0.916}  &\bf\textcolor{blue}{0.814} &0.814  &\bf\textcolor{blue}{0.925} & 0.920    \\
w/o FFM    &\bf\textcolor{blue}{0.913} &{0.915} &{0.811} &\bf\textcolor{blue}{0.817}  &0.921 &\bf\textcolor{blue}{0.928} \\ \hline
All &\bf\textcolor{red}{0.931} & \bf\textcolor{red}{0.937} & \bf\textcolor{red}{0.822} & \bf\textcolor{red}{0.828} &\bf\textcolor{red}{0.942} &\bf\textcolor{red}{0.945} \\  \bottomrule
\end{tabular}

\label{tab:ablation-2}
\vspace{-0.35cm}
\end{table}

\section{Experiment}
In this section, we first give the details of the experimental setup. Then we validate the proposed MD-VQA model with other mainstream VQA models on the proposed TaoLive database and the other two UGC compression VQA models. The ablation study and cross database validation are conducted to investigate the contributions of different groups of features and the generalization ability of the VQA models. Finally, we test the proposed MD-VQA model on two in-the-wild UGC VQA databases. 
\subsection{Benchmark Databases}
The proposed TaoLive database and two compressed UGC VQA databases including the LIVE-WC \cite{yu2021predicting} and YT-UGC$^+$ \cite{wang2021rich} databases are selected as the benchmark databases. For all the databases, we follow the common practice and spilt the databases with an 80\%-20\% train-test ratio. Additionally, all the databases are validated separately. To fully evaluate the stabilization and performance of the VQA models, the split is randomly conducted 30 times and the average results are recorded as the final performance.

\subsection{Implementation Details}
The EfficientNetV2 \cite{tan2021efficientnetv2} backbone is fixed with the EfficientNetV2-S weights pretrained on the ImageNet database \cite{deng2009imagenet} for semantic feature extraction while the ResNet3D-18 \cite{hara3dcnns} is fixed with the weights pretrained on the Kinetics-400 \cite{kay2017kinetics} database. All the frames are maintained with the original resolution for the semantic, distortion, and motion feature extraction. The Adam optimizer \cite{kingma2014adam} is employed with the initial learning rate set as 0.001. If the training loss has not decreased for 5 epochs, the learning rate will be reduced to half. The default number of epochs is set as 50.  The parameter $L$ described in Section \ref{sec:pre} is set as 8, which means 16 frames are uniformly sampled for the semantic and distortion feature extraction for a single clip.

\subsection{Competitors \& Criteria}
To fully evaluate the performance of the proposed method, we select several popular quality assessment models for comparison, which include BRISQUE\cite{brisque}, VSFA \cite{li2019quality}, TLVQM \cite{korhonen2019two}, VIDEVAL \cite{tu2021ugc}, PVQ \cite{ying2021patch}, BVQA \cite{li2022blindly} and SimpleVQA \cite{sun2022deep}. It's worth mentioning that BRISQUE belongs to NR-IQA models and we obtain the video quality features by averaging the features extracted from each frame with BRISQUE. 
The other VQA models are trained with the default parameter setting defined by their authors.

Two criteria are adopted to evaluate the performance of the quality assessment models, which include the Spearman Rank Order Correlation Coefficient (SRCC) and Pearson Linear Correlation Coefficient (PLCC). Before calculating the criteria values, a four-parameter logistic regression function \cite{sheikh2006statistical} is utilized to fit the predicted scores to the scale of MOSs. The value range for SRCC and PLCC is [0,1] and better models should yield higher SRCC and PLCC values.

\subsection{Performance Discussion}
The experimental performance on the three compressed UGC VQA databases is shown in Table \ref{tab:intra}, from which we can draw several conclusions. (a) The proposed MD-VQA achieves first place and surpasses the second place (SimpleVQA \cite{sun2022deep}) by about 0.004, 0.033, and 0.010 in terms of SRCC values on the LIVE-WC, YT-UGC+, and TaoLive databases respectively, which demonstrates its effectiveness of predicting the quality levels of compressed UGC videos. (b) The handcrafted-based methods (BRISQUE, TLVQM, and VIDEVAL) are significantly inferior to the deep learning-based methods (VSFA, PVQ, BVQA, SimpleVQA, and MD-VQA). It can be explained that the hand-crafted based methods hold the prior experience of NSS, which comes from the pristine videos. However, the characteristics of compressed UGC videos are far more complicated and do not suit the prior knowledge of natural regularities. (c) All the VQA methods experience performance drops on the YT-UGC$^+$ database compared with the other two databases. The YT-UGC$^+$ database uses the recommended VP9 settings and target bit rates \cite{vp9} for compression while the LIVE-WC and TaoLive databases control the compression by varying the CRF parameters of H.264 and H.265 respectively. It might be because the recommended VP9 compression settings 
do not monotonically reduce the video bit rates, thus being more challenging for quality prediction.

\subsection{Ablation Study}
To investigate the contributions of different features employed in MD-VQA, we conduct the ablation study in this section. The experimental results for employing different types of features are shown in Table \ref{tab:ablation}. Combining features yield better performance than using a single group of features and employing all features leads to the best performance among the combinations of different features, which confirms the contributions of the semantic, distortion, and motion features. Additionally, by comparing the performance of SF, DF, and MF models, we can see that the SF model achieves first place on all the databases, which indicates that the semantic features make the most devotion to the final performance. What's more, the distortion and motion features achieve only 0.277 and 0.537 in terms of SRCC values on the YT-UGC$^+$ database. This is because 1/3 of the YT-UGC$^+$ database's compressed UGC videos are obtained from game videos. The game videos' distortion and motion characteristics differ greatly from the natural videos. To illustrate, noise usually does not exist in game videos and the motion blur effect is manually introduced if applied. Such phenomena result in the relatively low performance of the distortion and motion features. 
We also conduct the ablation study for using absolute error (ABS) and feature fusion module (FFM) and the results are listed in Tab \ref{tab:ablation-2}, from which we can find that both ABS and FFM make contributions to the final results.

\subsection{Cross Database Performance}
\label{sec:cross}
Since UGC videos are diverse in contents and distortions, We carry out the cross database validation to test the generalization ability of the VQA models in this section. The VQA models are trained on the TaoLive database and tested on the other two compressed UGC VQA databases. The experimental results are listed in Table \ref{tab:cross}. The proposed MD-VQA model has surpassed all the compared VQA models on both LIVE-WC and YT-UGC$^{+}$ databases, which proves its strong generalization ability. The handcrafted-based VQA models (BRISQUE, TLVQM, and VIDEVAL) perform badly on the YT-UGC$^{+}$ database and the deep learning-based methods also undergo significant performance drops from the LIVE-WC database to the YT-UGC$^+$ database. The reason is that the YT-UGC$^+$ database employs VP9 \cite{vp9} while the TaoLive database utilizes H.265 for compression respectively. The different compression standards can bring unignorable gaps between the data distributions. Therefore the quality representation learned from the TaoLive database is less effective on the YT-UGC$^+$ database but works well on the H.264 compressing LIVE-WC database.

\begin{table}[!th]\small
\centering
\caption{Experimental performance of cross databases, where the VQA models are all pretrained on the TaoLive database. }
\setlength{\tabcolsep}{4.5pt}
\begin{tabular}{c|cc|cc}
\toprule
\multirow{2}{*}{Method} & \multicolumn{2}{c|}{LIVE-WC} & \multicolumn{2}{c}{YT-UGC$^+$} \\  \cline{2-5}
         & SRCC$\uparrow$      & PLCC$\uparrow$     & SRCC$\uparrow$     & PLCC$\uparrow$ \\ \hline

BRISQUE\cite{brisque}   &0.708 &0.709 &0.026 &0.059 \\
TLVQM\cite{korhonen2019two}  &0.562 &0.583 &0.155 & 0.184      \\
VIDEVAL\cite{tu2021ugc}     &0.557 &0.583 &0.077 &0.132         \\
VSFA\cite{li2019quality}    &0.701 &0.698 & 0.357 & 0.399  \\
SimpleVQA\cite{sun2022deep}  &\bf\textcolor{blue}{0.711} &\bf\textcolor{blue}{0.723} &\bf\textcolor{blue}{0.388} &\bf\textcolor{blue}{0.394}\\ \hline

MD-VQA   &\bf\textcolor{red}{0.742} & \bf\textcolor{red}{0.728} &\bf\textcolor{red}{0.440} &\bf\textcolor{red}{0.448} \\

                      \bottomrule
\end{tabular}

\label{tab:cross}
\vspace{-0.3cm}
\end{table}

\begin{table}[!th]\small
\centering
\caption{Experimental performance of in-the-wild UGC databases. }
\setlength{\tabcolsep}{4pt}
\begin{tabular}{c|cc|cc}
\toprule
\multirow{2}{*}{Method} & \multicolumn{2}{c|}{KonViD-1k} & \multicolumn{2}{c}{LIVE-VQC}  \\  \cline{2-5}
& SRCC$\uparrow$       & PLCC$\uparrow$      & SRCC$\uparrow$       & PLCC$\uparrow$\\\hline

BRISQUE\cite{brisque} & 0.657   & 0.658    & 0.593   & 0.638         \\
TLVQM\cite{korhonen2019two}  & 0.773   & 0.769    & 0.799    & 0.803             \\
VIDEVAL\cite{tu2021ugc}  & 0.783   & 0.780    & 0.752    & 0.751               \\
VSFA\cite{li2019quality} & 0.785   & 0.797    & 0.716    & 0.775           \\
SimpleVQA\cite{sun2022deep}   &\bf\textcolor{red}{0.856} &\bf\textcolor{red}{0.860} &\bf\textcolor{blue}{0.811} &\bf\textcolor{blue}{0.815}              \\ \hline

MD-VQA    & \bf\textcolor{blue}{0.851}   & \bf\textcolor{blue}{0.853}   & \bf\textcolor{red}{0.814} & \bf\textcolor{red}{0.839}   \\

                      \bottomrule
\end{tabular}

\label{tab:ugc}
\vspace{-0.2cm}
\end{table}

\subsection{In-the-wild Performance}
Although the proposed MD-VQA model focuses on the compressed UGC VQA issues, we also test its performance on some mainstream in-the-wild UGC VQA databases, which includes the KoNViD-1k\cite{hosu2017konstanz} and LIVE-VQC\cite{sinno2018large} databases. These databases are not focused on compression and contain a wide range of distortions. Validation on these databases can reflect the VQA models' ability to handle general UGC VQA issues. The experimental results are exhibited in Table \ref{tab:ugc}, from which we can find that the proposed MD-VQA outperforms the compared VQA models on the LIVE-VQC databases and gets the second-ranking on the KoNViD-1k database. This implies that the proposed MD-VQA remains competitive not only for compression-specific VQA issues and can be taken as a strong baseline for in-the-wild UGC VQA tasks as well.



\section{Conclusion}
In this paper, we focus on the compressed UGC VQA issues. To meet the practical needs of live platforms, we carry out a large-scale compressed UGC VQA database called \textbf{TaoLive}. Unlike the common compression VQA databases that employ high-quality videos as source videos, the TaoLive database collects the 418 source UGC videos that cover a wide quality range and generate the compressed videos by varying the CRF parameters with H.265. A well-controlled subjective experiment is conducted to gather the quality labels for the compressed UGC videos. Further, we propose a VQA model (MD-VQA) to assess the quality of the compressed UGC videos from the semantic, distortion, and motion dimensions. The extensive experimental results confirm the effectiveness of the proposed method.

\section{Acknowledgement}
This work was supported in part by NSFC (No.62225112, No.61831015), the Fundamental Research Funds for the Central Universities, National Key R\&D Program of China 2021YFE0206700, and Shanghai Municipal Science and Technology Major Project (2021SHZDZX0102).

{\small
\bibliographystyle{ieee_fullname}
\bibliography{egbib}
}

\end{document}